\documentclass[conference]{IEEEtran}
\pdfoutput=1
\IEEEoverridecommandlockouts
\usepackage{cite}
\usepackage{amsmath,amssymb,amsfonts}
\usepackage{algorithmic}
\usepackage{graphicx}
\usepackage{textcomp}
\usepackage{xcolor}
\usepackage[misc]{ifsym} 
\usepackage{threeparttable}
\def\BibTeX{{\rm B\kern-.05em{\sc i\kern-.025em b}\kern-.08em
    T\kern-.1667em\lower.7ex\hbox{E}\kern-.125emX}}
\begin{document}

\title{IDEA: Interactive Double Attentions from Label Embedding for Text Classification *}


\author{\IEEEauthorblockN{Ziyuan Wang, Hailiang Huang, Songqiao Han$^{(\textrm{\Letter})}$ \thanks{Songqiao Han$^{(\textrm{\Letter})}$ is the corresponding author.}} \thanks{*The article has been accepeted by ICTAI2022}
\IEEEauthorblockA{\textit{AI Lab, School of Information Management and Engineering} \\
\textit{Shanghai University of Finance and Economics, China} \\
wangziyuan@163.sufe.edu.cn, \{hlhuang, han.songqiao\}@shufe.edu.cn}
}

\maketitle

\begin{abstract}
Current text classification methods typically encode the text merely into embedding before a naive or complicated classifier, which ignores the suggestive information contained in the label text. As a matter of fact, humans classify documents primarily based on the semantic meaning of the subcategories. We propose a novel model structure via siamese BERT and interactive double attentions named IDEA ( \textit{I}nteractive \textit{D}oubl\textit{E} \textit{A}ttentions) to capture the information exchange of text and label names. Interactive double attentions enable the model to exploit the inter-class and intra-class information from coarse to fine, which involves distinguishing among all labels and matching the semantical subclasses of ground truth labels. Our proposed method outperforms the state-of-the-art methods using label texts significantly with more stable results.
\end{abstract}


\section{Introduction}
Recently, BERT\cite{devlin2019bert} and its variants have achieved great success in various NLP tasks via dynamic text representation, such as supervised text classification. However, the traditional paradigm usually encodes text embeddings merely without label information, lacking of key information for downstream classifier to contrast.

Considering label information has recently emerged as a promising research direction. Designing one-side attention mechanisms \cite{wang_joint_2018,sun_hierarchical_2019,wang_concept-based_2021}, learning distribution of predictions with ground truth labels \cite{guo_label_2021}, joining documents and label words \cite{xiong_fusing_2021} are main techniques in exploiting labels information. Meanwhile, label words also play an important role in an unsupervised situation \cite{meng_text_2020,shen_taxoclass_2021}, few-shot \cite{pappas_gile_2019,halder_task-aware_2020,luo_dont_2021}, zero-shot \cite{nam_all-text_2016} and multi-label tasks \cite{xiao_label-specific_2019,zhang_correlation-guided_2021}. None of these previous works consider the interactive attention between the text and the labels, which limit the classification performance.

Nowadays, interactive attention has been used widely in several NLP tasks. BIDAF \cite{seo_bidirectional_2018} combines the answer-word attention and document in reading comprehension. Similarly, IAN \cite{ma_interactive_2017} encodes targets and contexts separately in aspect-level sentiment classification, then CIA \cite{Yu_2017_ICCV} and DSACA \cite{liu_dual_coatt_2022} employ a multi-modal attention involving text and visual features in question answering.

Motivated by these observations and inspired by interactive attention learning \cite{ma_interactive_2017, seo_bidirectional_2018}, we propose a novel approach that builds a query-aware context representation processing, making using of the interactive information to judge categories, namely \textbf{I}nteractive \textbf{D}oubl\textbf{E} \textbf{A}ttention(\textbf{IDEA}). IDEA incorporates the label interactive features into sentence embeddings while fusing text interactive features into label embeddings from double aspects. The main purpose of IDEA is to offer the model more assitance source especially when the label text is informative for classification. (When the label text is random or inaccurate, we supposing it will makes the model more robust.)

As the illustrated shown in Fig. 1, an original text from AGNews dataset, for instance, there are several semantic subclasses within category "Sports" or "Business", including "basketball", "football", "Accounting", "Marketing" etc. Obviously, tokens in text such as "plans, buy, water firm" reveal the core "Business" related semantic meaning. But certain tokens like "Lakers, basketball, Vlade Divac" are strongly related to "Sports", which may mislead the classifier into choosing the ground truth. As the blue dashed lines indicate, the semantically related subclass from ground truth like "Accounting" should have a higher similarity to the context.

\begin{figure}[!htb]
  \centering
  \includegraphics[width=1.1\linewidth]{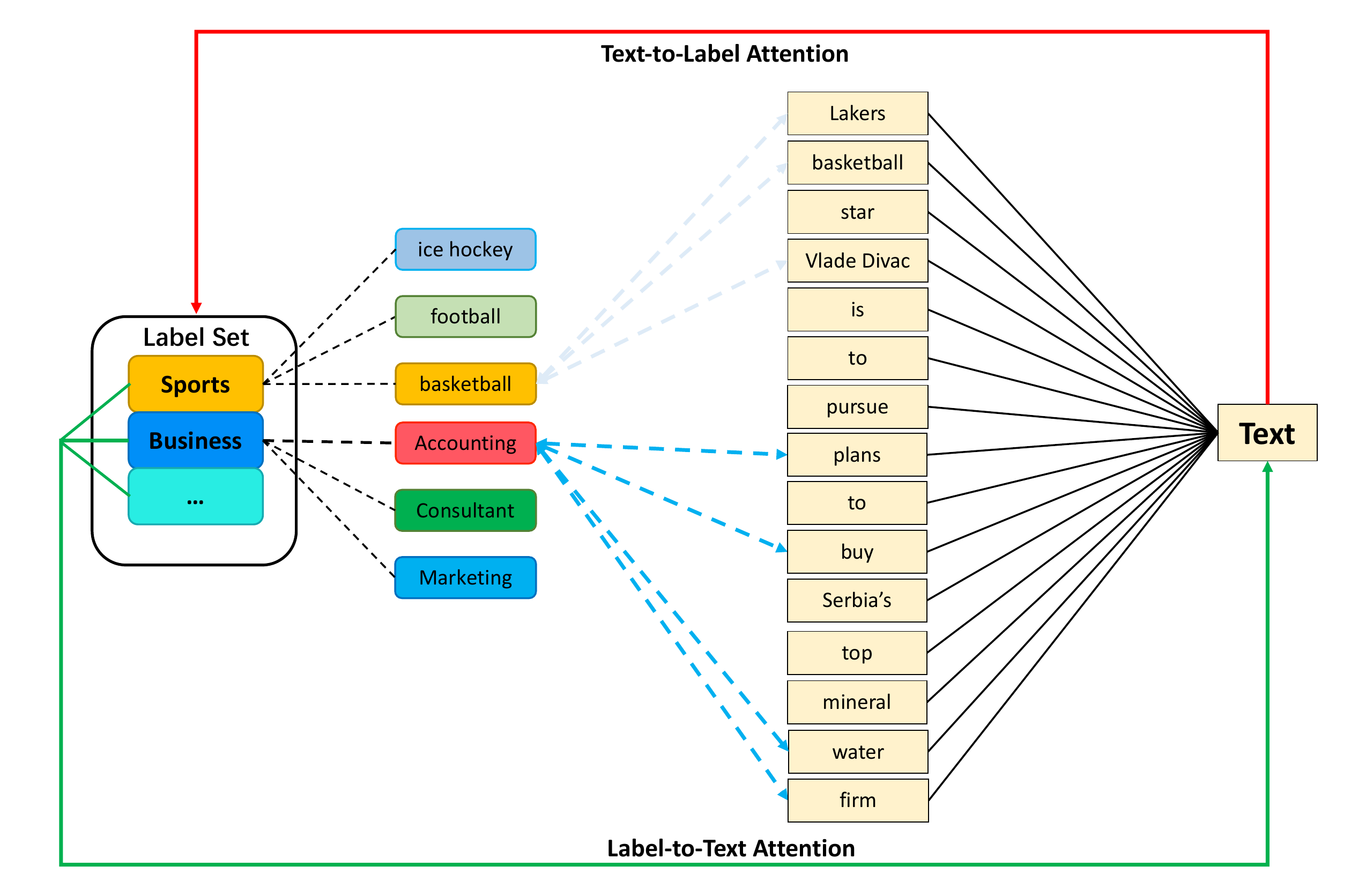}
  \caption{Overview of IDEA illustration. Original text from AGNews training dataset-20815. Ground truth: \textbf{Business} instead of Sports}
\end{figure}

Consequently, we assume the red line where text-token attention fuses label set global information will broaden the whole label space for the model to distinguish explicitly the \textbf{inter-class} relations from a coarse-grained overall perspective, thereby making the model easier to distinguish the ground truth based on the entire label set options.

In terms of labels, the green line where label-token attention fusing text global information reveals \textbf{intra-class} correlations implicitly between semantical subclass and its category, from a fine-grained partial perspective. Owing to the prominent character of BERT discovering hypernyms and hyponyms \cite{hanna-marecek-2021-analyzing, beloucif2021how, ravichander-2020-systematicity}, the higher text-to-label attention can match the related tokens "plans, buy, firm" with ground truth "Business" via semantical subclass "Accounting", namely its hyponym. In contrast, "Lakers, basketball" have lower attention scores for the deviation from core theme.

In conclusion, our approach build a connection to match the category-related words in the document with the semantical-related subclasses of ground truth labels through the interactive double attentions from coarse to fine. We summarize our contributions as following: \textit{(1) Siamese View.} Beginning with the siamese BERT architecture as a backbone, we can derive semantically meaningful sentence embeddings of labels and documents in the same BERT latent space. Empirically, siamese network works well with off-the-shelf architectures as a basic global embedding strategy\cite{jing2022masked}. \textit{(2) Interaction View.} In contrast to other methods utilizing labels, the cross interactions couple the label and context vectors , complementing each other to perform the classification together.

\section{Methodology}

\subsection{Problem Formulations}
Let $X = [x_1,\cdots,x_j,\cdots,x_N]\in\mathbb{R}^{1\times N}$ denote one document consisting of $N$ words. Each document has one label contained in $[y_1,\cdots,y_i,\cdots,y_L]\in\mathbb{R}^{1\times L}$, where $L$ is the number of class, $y_i$ is the label word. Each sentence word and label word can be encoded a sequence of words into a high-dimensional space and represented as a $d$-dimension vector via BERT model.

\subsection{IDEA Model Structure}
Our proposed method \textbf{I}nterac{}tive \textbf{D}oubl\textbf{E} \textbf{A}ttention (IDEA) is given in Figure 2. 
\begin{figure}[!htbp]
  \centering
  \includegraphics[width=1.05\linewidth]{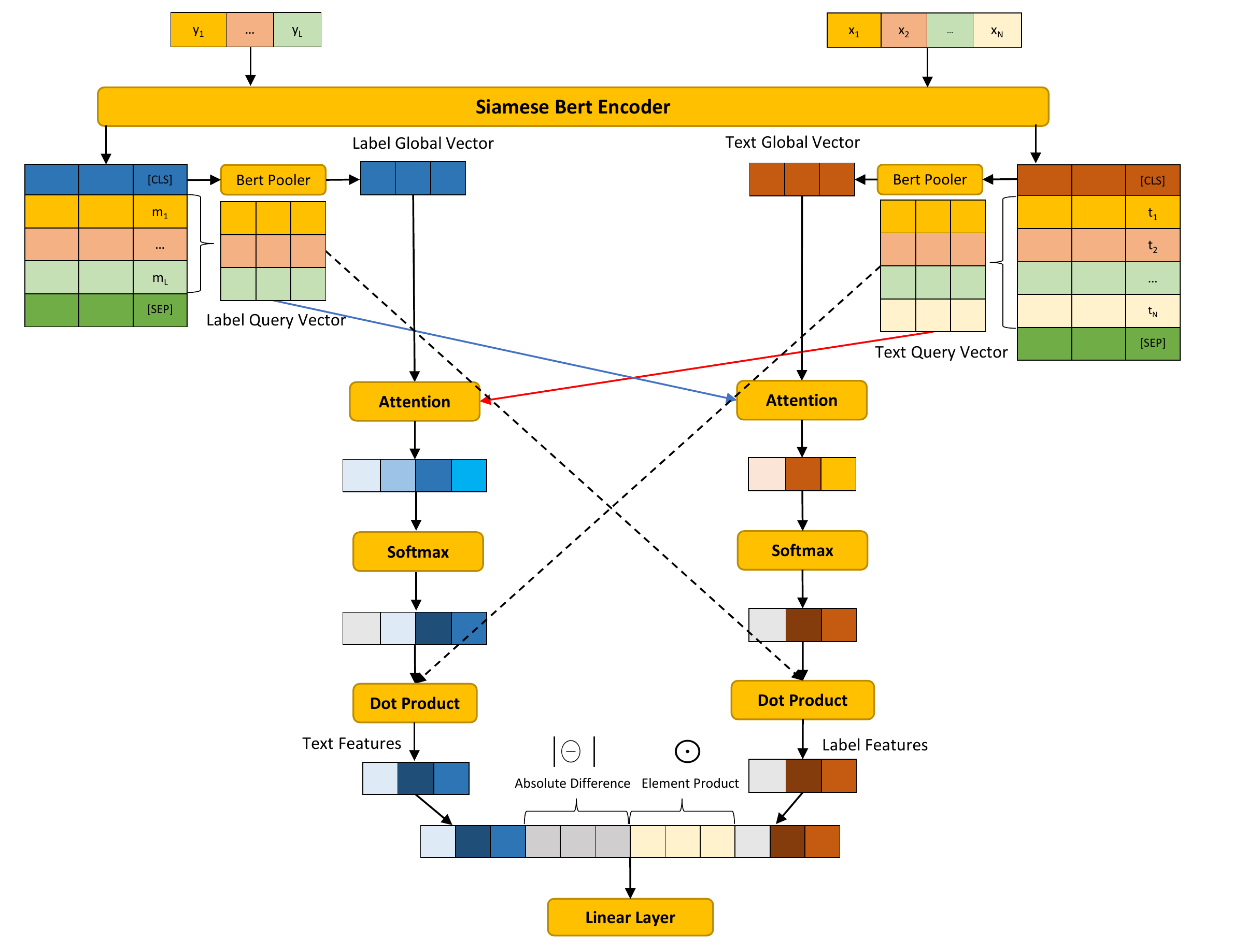}
  \caption{Overview the Architecture of IDEA. The red, blue solid arrows and dash lines all represent the interactions between labels and texts through Attention or matrix multiplication.}
\end{figure}

\subsubsection{Text Representation}~\\
At first, we input text and label into BERT Encoder to derive text vectorized representation. As is known to us, output vector of [CLS] token is produced by \textit{BERT Pooler Layer}, which is a weighted added by all token-embeddings through a Dense Layer. In general, we consider [CLS] token as the global representation of a whole sentence and make use of the embedding of it to downstream classifier.

Considering within a batch $D=[X_1,\cdots,X_k,\cdots,X_K]$ including $K$ texts, we suppose the output text embedding after BERT encoder is $[T_1,\cdots,T_k,\cdots,T_K]$, and 
$$T_k = [t_{k,\text{[CLS]}},t_{k,1},\cdots,t_{k,j},\cdots,t_{k,N},t_{k,\text{[SEP]}}]$$, where $t_{k,j}\in\mathbb{R}^{d},T_k \in\mathbb{R}^{K\times(N+2)\times d}$. In the same way, the label embedding is $$M_k = [m_{k,\text{[CLS]}},m_{k,1},\cdots,m_{k,i},\cdots,m_{k,L},m_{k,\text{[SEP]}}]$$, where $m_{k,i}\in\mathbb{R}^{d}, M_k \in\mathbb{R}^{K\times(L+2)\times d}$. Then we define the text-token embedding $$[t_{k,1},\cdots,t_{k,j},\cdots,t_{k,N}]\in\mathbb{R}^{K\times N\times d}$$ as text query vectors, $t_{k,\text{[CLS]}}$ as text global vectors $t_{k,\text{global}}$, and the label-token embedding $$[m_{k,1},\cdots,m_{k,i},\cdots,m_{k,L}]\in\mathbb{R}^{K\times L\times d}$$ as label query vectors, $m_{k,\text{[CLS]}}$ as label global vector $m_{k,\text{global}}$.

\subsubsection{Interavtive Attention}~\\
\textbf{Text Attention}. Based on the assumption that global label information is suggestive in selecting high-correlation tokens in a coarse-grained way, the attention mechanism is designed to derive the query-aware context representation. We suppose each token in the input text as a query to pay more attention to the category-related words. 
Firstly we utilize a score function $$t_{k,j}^{\prime} = \tanh(t_{k,j}\cdot W_m\cdot m^T_{k,\text{global}}+b_m)$$, where $ W_m\in\mathbb{R}^{d\times d}, b_m\in\mathbb{R}^{K\times N}$ by fusing label global representation, then we can derive text attention vector via softmax function $\alpha_{k,j}=\frac{\exp{(t_{k,j}^{\prime})}}{\sum\exp{(t_{k,j}^{\prime})}}\in\mathbb{R}^{K\times N}$. Then text freatures are equal to multiplying text attention by the text query vectors, i.e. $$c_k = \sum_j\alpha_{k,j}\cdot t_{k,j}, \text{where} \quad \boldsymbol{c}\in \mathbb{R}^{K\times d} $$\\
\textbf{Label Attention}. In parallel, the text global information can represent the whole sentence, which can dig out the fine-grained semantic meaning of one label word. There exist some hidden semantical syntactical features of every label word token embedding. 
As before, we can derive $$m_{k,i}^{\prime} = \tanh(m_{k,i}\cdot W_t\cdot t^T_{k,\text{global}}+b_t)$$, where $W_t\in\mathbb{R}^{d\times d}, b_t\in\mathbb{R}^{K \times L}$ by fusing text global representation, we can derive label attention vector via softmax function $\beta_{k,i}=\frac{\exp{(t_{k,i}^{\prime})}}{\sum\exp{(t_{k,i}^{\prime})}}\in\mathbb{R}^{K\times L}$. Identically, the label features is $$s_k = \sum_i \beta_{k,i} \cdot m_{k,i}, \text{where} \quad \boldsymbol{s} \in\mathbb{R}^{K\times d}$$

\subsubsection{Weighted Similarity Features}~\\
Furtherly, in order to explore the similarity between the interactive double attentions features, we add some features as new similarity measurement. The first is the element-wise product, the other is element-wise absolute difference \cite{conneau2017supervised,sun_hierarchical_2019}.

The element-wise product can be defined as $p_k = c_k \odot s_k\in\mathbb{R}^{K\times d}$, where $\odot$ means Hadamard product, i.e. element-wise product. And the element-wise absolute difference is $d_k = |c_k\ominus s_k|\in\mathbb{R}^{K\times d}$, where $\ominus$ means element-wise minus,$|\cdot|$ means absolute value.

After computing two kinds of similarity features, we can derive their weights based on the attention vectors $p_k$ and $d_k$ by:

Weighted Absolute Element-wise Difference can be defined as: 
$$p_k^{\prime} = \gamma_k p_k, \text{where}\quad\gamma_k = \frac{\exp(\mu(\sum_k d_k))}{\exp(\mu(\sum_k p_k))+\exp(\mu(\sum_k d_k)}$$,
and $\mu(\cdot)$ is the average function.

Following aforementioned setting, we choose the probability-like way to obtain the weighted element-wise product:
$$d_k^{\prime} = \eta_k d_k, \text{where}\quad\eta_k = 1-\gamma_k$$

Finally, the final embeddings are concatenated as a vector $\boldsymbol{z}$ fed into a downstream classifier:
$$\boldsymbol{z} = [\boldsymbol{c} \oplus \boldsymbol{p^{\prime}} \oplus \boldsymbol{d^{\prime}} \oplus \boldsymbol{s}] \in\mathbb{R}^{K\times4d}$$,
where $\oplus$ means concatenate operation.

\subsubsection{Linear Classifier and Loss}~\\
Lastly, we place a linear layer $f(\cdot)$ to project logits dimension $d$ into the space of the targeted $L$ classes:
$$\boldsymbol{\hat{y}} = \text{softmax}(f(\boldsymbol{z})), \text{where}\quad\boldsymbol{\hat{y}}\in\mathbb{R}^{K\times L}$$
Based on the predicted labels and ground truth labels, the loss can be calculated with L2 regularization as:
$$\mathcal{L} = -\frac{1}{K} \sum_{i=1}^L \sum_{k=1}^{K} y_{k,i} \log \left(\hat{y}_{k,i}\right) + \frac{\lambda}{2} \|\boldsymbol{W}\|_{F}^2$$
, where $\lambda>0$ is a tuning parameter. In order to minimize the loss function, we train the model end-to-end under the settled parameters.

\section{Experiments}

\subsection{Datasets}
To evaluate the effectiveness of IDEA model, we choose 5 benchmark datasets by comparing with the state-of-the-art methods using label texts. All the datasets we used were originally constructed by Zhang et al.(2015)\cite{zhang_CNN_2015}

\begin{itemize}
  \item \textbf{AG News}. A news article dataset with titles and descriptions is constructed by choosing 4 classes: World, Sports, Business, and Sci\&Tech, containing 120,000 training samples and 7600 for testing.
  \item \textbf{DBpedia}. An ontology classification with titles and contents containing 560,000 samples for training and 70,000 for testing  over 14 ontology classes including Company, Artist, and so on.
  \item \textbf{Yahoo! Answers Topic}. A questions\&answering dataset with question title, question content and best answer containing 1,400,000 training samples and 60,000 testing samples with ten categories including Society\&Culture, Health, Sports, etc.
  \item \textbf{Yelp Review Full}. A dataset extracted from Yelp Dataset Challenge 2015 data containing 650,000 training samples and 50,000 testing samples for each review star from 1 to 5. We replace the star with classes: ['bad', 'poor', 'fair', 'good', 'excellent'].
  \item \textbf{Yelp Review Polarity}. A dataset also extracted from Yelp Dataset Challenge 2015 data but coarsely divided with two classes, considering 1 and 2 stars as negative, and 4 and 5 as positive, containing 560,000 training samples and 38,000 testing samples. We replace the star with classes: [ 'negative', 'positive'].

\end{itemize}
Since the original benchmarks do not include the development dataset, we randomly stratified sampling the validation dataset whose size is equal to test dataset from the original train dataset.

Given the enormous size of the Yahoo dataset, we choose a subset of size 5\% samples (i.e. 70,000 for training samples and 3,000 for test samples) in our experiments for considering the computational resources.

\begin{table*}[ht!] \label{tab:number}
\centering
\begin{threeparttable}
  
  \caption{Accuracy(\%) on Different Test Datasets\tnote{$\dagger$}}
  \begin{tabular}{c|cccccc} 
  \textbf{Models} & \textbf{AGNews}         & \textbf{DBpedia}        & \textbf{Yahoo}          & \textbf{YelpF}          & \textbf{YelpP}           & \textbf{avg}      \\ 
  \hline
  LEAM            & 92.45                  & 99.02                  & \textbackslash{}        & 64.09                  & 95.31                   & \textbackslash{}  \\
  BERT            & 94.17±0.75*         & 99.18±0.02          & 72.77±1.51**        & 67.22±0.19          & 95.62±0.02*          & 85.79           \\
  LSAN            & 88.43±6.24**        & \textbf{99.26±0.04}          & 60.90±1.73***        & 62.45±0.19**        & 96.41±0.97*          & 81.49           \\
  FLEBw.[SEP]     & 94.18±0.55**       & 99.19±00.02 & 72.56±1.24***       & 67.20±0.08          & 95.63±0.06*          & 85.75           \\
  FLEBw.o.[SEP]   & 94.57±0.18***       & 99.18±0.02          & 72.95±1.32***       & 67.24±0.2          & 97.51±0.03*          & 86.29           \\ 
  \hline
  IDEA            & \textbf{94.92±0.11} & 99.18±0.02          & \textbf{75.15±0.41} & \textbf{67.25±0.27} & \textbf{97.52±0.06}~ & \textbf{86.81}  \\
\end{tabular}

\begin{tablenotes}
\footnotesize
\item[$\dagger$] Statistical significant difference at p-value $<$ 0.01, 0.05, and 0.1 are denoted as ***, **, * respectively in Welch's T-test(unequal variances). The bold type of every column means the best result within each dataset.
\end{tablenotes}
\end{threeparttable}
\end{table*}

\subsection{Baseline Models}
All approaches for comparison are listed below:
\begin{itemize}
  \item \textbf{LEAM}\cite{wang_joint_2018} (Wang et al., 2018) came up with an attention mechanism are used to jointly learn the embedding of words and labels via Word2Vec.
  \item \textbf{BERT}\cite{devlin2019bert} (Delvin et al., 2019) is the classic pretrained language representation model that generates contextualized dynamic word vectors.
  \item \textbf{LSAN}\cite{xiao_label-specific_2019} (Xiao et al., 2019) applies a label-specific attention network model based on self-attention and label-attention mechanism. Here, we provide a comparable baseline method LSAN$_\text{BERT}$ , which uses BERT to provide text token encoding, but its labels are encoded independently.
  \item \textbf{FLEBw.[SEP]}\cite{xiong_fusing_2021} (Xiong et al., 2021) is \textbf{F}using \textbf{L}abel \textbf{E}mbedding into \textbf{B}ert, a method that concatenates document words and labels texts with [SEP] token. Since the authors haven't provided the source code yet, we implement the model by ourselves based on the paper.
  \item \textbf{FLEBw.o.[SEP]}\cite{xiong_fusing_2021} (Xiong et al., 2021) is also a method that concatenate document words and labels texts without [SEP] token, which we also implement ourselves.
\end{itemize}
We reproduce all the baseline experiments ourselves with the same experiment settings except LEAM, for its specialized architecture for Word2Vec. Most of the baseline results are close to the original results and some are even better.

\subsection{Experiment Settings}
In our experiments, all word embeddings from context and labels are encoded by BERT\cite{devlin2019bert}(bert-base-uncased\footnote{https://huggingface.co/bert-base-uncased}).

We choose the batch size of 32 by means of dynamic padding. There's no need to pad all text to a fixed size decided by max length in whole documents but only guaranteed the same length within a training step, which will save computational resources. Besides, we set the learning rate to $5\times10^{-5}$, and the coefficient of L2 regularization in the objective function is set to $0.01$, with the dropout rate of 0.1. In order to optimize the parameters of IDEA effectively, we employ the AdamW \cite{loshchilov2018decoupled} with epsilon of $1\times10^{-6}$ to update vector.

According to the evaluation result, we choose the best epoch as 2/3/2/5/5 with the highest accuracy in validation datasets for AGNews, DBpedia, Yahoo, Yelpp, Yelpf separately. In the end, we report those models’ performance on the test set in terms of accuracy as evaluation metrics. Meanwhile, we set the same model structure parameter as \cite{xiao_label-specific_2019} in baseline LSAN experiments.

Label embeddings are encoded by the default given order of dataset, with commas used to separate multiple tokens of a single label, such as "company, educational institution, artist, athlete, office holder, $\cdots$".

\subsection{Main Results}
\textbf{Our method returns significantly better average results over five datasets than baselines.}
For text classification task, as shown in Table 1, our method significantly outperforms the best baseline, namely, FLEBw.o.[SEP], whose conclusion is consistent with FLEB\cite{xiong_fusing_2021}. Firstly, IDEA gains an obvious boost in Yahoo dataset. Besides, our method also outperforms all baselines in the separate dataset in terms of accuracy except DBPedia. Due to the simplicity of DBPedia as we speculated, the IDEA cannot provide effective information, the result of LSAN is slightly better than IDEA, however it's not statistically significantly better than IDEA. As a result, we consider that IDEA can explore the complex semantic similarity between texts and labels, leading to competitive results. Due to sampling subset of data instead of the original Yahoo dataset so we don't display result of LEAM at Yahoo here. In contrast, IDEA is more expert at dealing with "difficult to distinguish" texts.

\textbf{The results reveal that IDEA is more stable on different random seeds.} In Table 1, IDEA obtains more stable results with a smaller standard deviation compared to the majorities. In contrast, the LSAN \cite{xiao_label-specific_2019} performs the best result at DBpedia dataset. However, LSAN is more unstable with a larger standard deviation on different random seeds. Considering the difference of the model structure, LSAN only builds one-side label attention and self-attention instead of double-side attention. In AGNews group, FLEBw.o.[SEP] has a higher value but a lower p-value compared with BERT. According to Welch's T-test $t=(\mu_a-\mu_{\text{IDEA}})/(\sqrt{S_a^2/n_a+S_{\text{IDEA}}^2/n_{\text{IDEA}}})$, the higher mean and the lower standard deviation FLEBw.o.[SEP] derives a lower p-value, leading to this counterintuitive result. As a result, we hold on view that double attentions entirely complement the hidden semantic information, guranteed the robustness of IDEA.

\begin{figure}[!htb]
  \centering
  \includegraphics[width=1.0\linewidth]{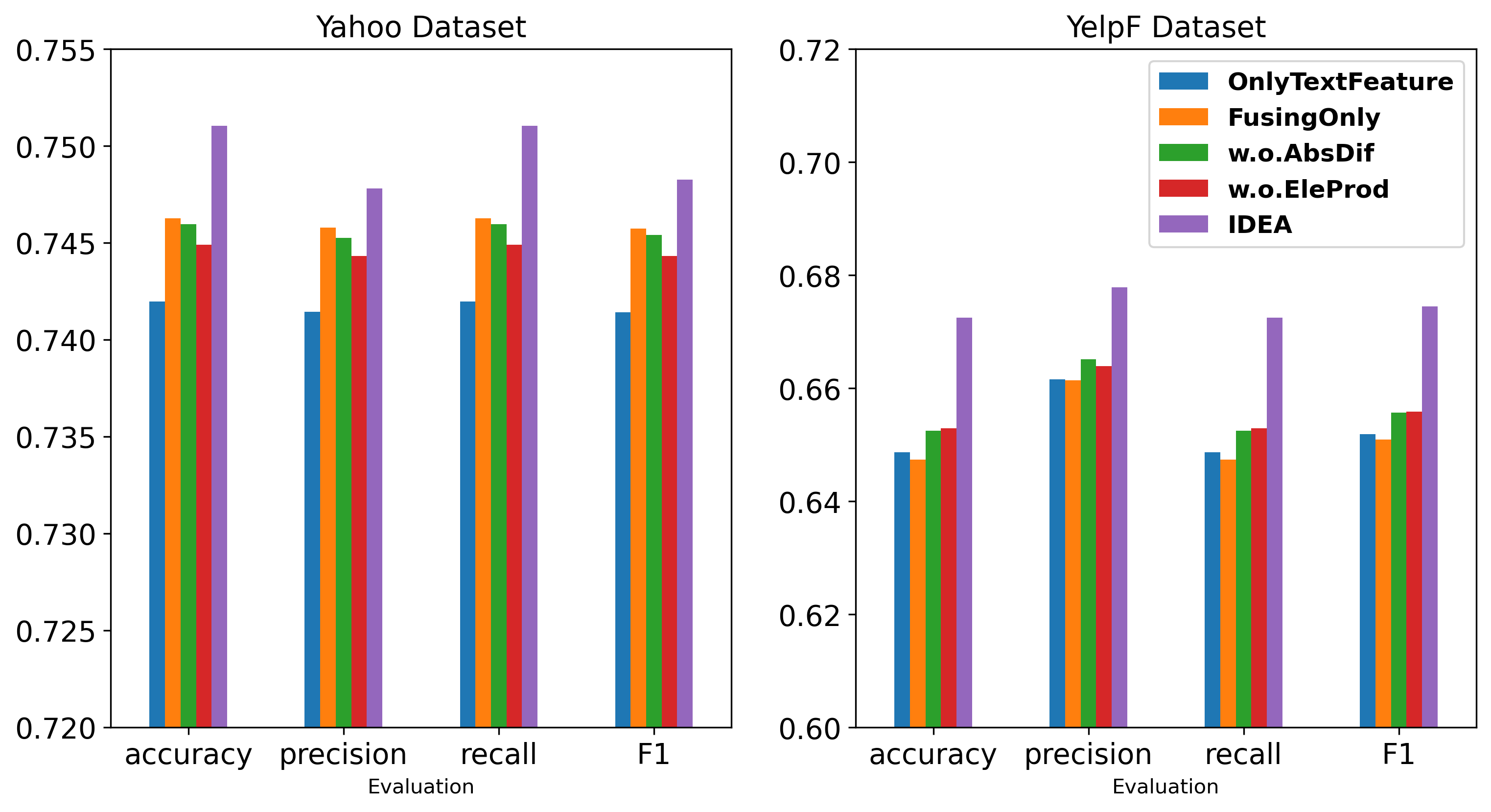}
  \caption{Evaluations Results of Ablation Tests on Yahoo and YelpF datasets. The purple columns are the evaluation results of IDEA. All the average results of 5 times experiments are statistically significant with p-value$<$0.05.}
\end{figure}

\subsection{Ablation Study}
It's obvious that text features play a crucial and essential role in representing the entire sentence. In order to explore the impact of different components of the IDEA model on effectiveness, we design different groups of controlled experiments in terms of 'OnlyTextFeatures', 'OnlyFusing' (without label\&text features), 'w.o.AbsDif' (without Element Absolute Difference features), 'w.o.EleProd' (without Element Product features) on two relatively difficult datasets.

As we can observe in Figure 3, IDEA greatly enhances overall evaluations in various aspects. In particular, as we can see in Yahoo dataset, label attention contributes greatly to recall value with an over 1\% rise in recall value compared with 'OnlyTextFeature', which can prove the validity of itself. The steep reduction of the accuracy is also significant when removing two elements' similarity features. Hence, it is no wonder that combining all the factors achieves the best performance on all the experimental datasets.

Last but not least, we visualize the learned representation of final output features of IDEA via t-SNE\cite{tsne} in Figure 4. There exist separate and distant hyperplanes in vector space that will support classification effectively.

\begin{figure}[!htb]
  \centering
  \includegraphics[width=1.0\linewidth]{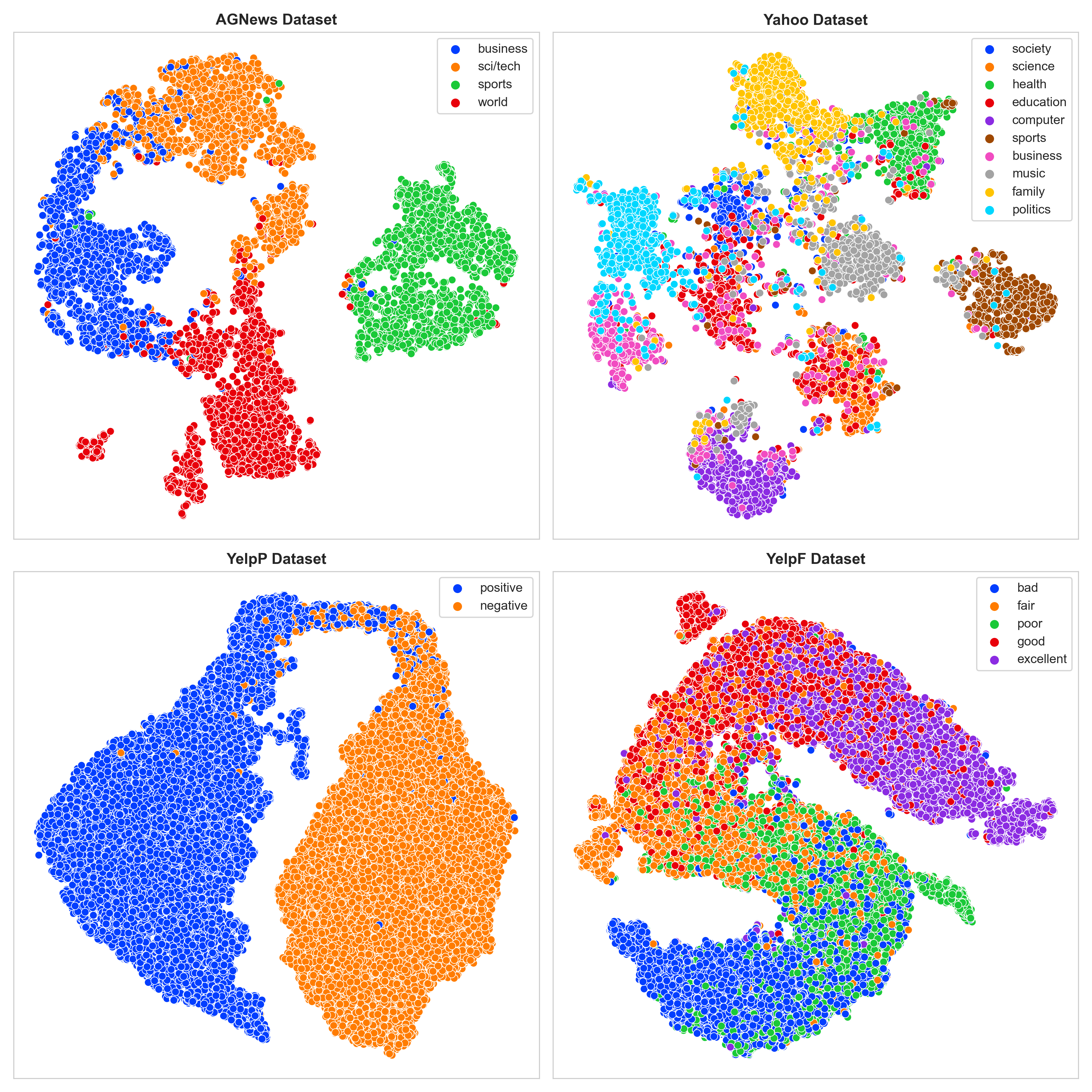}
  \caption{t-SNE visualization results of final features over four datasets IDEA raise performance.}
\end{figure}

\section{Conclusion}
In this work, we propose IDEA, an interactive double attentions approach, to explore the label information better to enhance text classification results. Specifically, the IDEA model utilizes label-to-text and text-to-label attention to effectively leverage the cross-sentence interaction from overall to partial, connecting category-related words with the semantical-related subclasses of groud truth labels. Experiments over five text classification datasets show that IDEA achieves consistent and significant gains and outperform five benchmarks. We will explore the circumstance that the label's text has poorly semantic association with texts. In addition, we intend to further investigate hypernyms and hyponyms of category words using IDEA in future work.

\section{Acknowledgements}
This work was supported by the National Natural Science
Foundation of China (under Project No. 72271151), the
graduate innovation fund of Shanghai University of Finance
and Economics (under Project No. CXJJ-2021-351), and the China Scholarship Council (202106480015).

\bibliographystyle{IEEEtran.bst}
\bibliography{IEEEabrv, IDEA_arxiv.bib}
\end{document}